# Predicting Euler Characteristics and Constructing Topological Structure Using Machine Learning Techniques

Gyunghun Yu[1,], Seong Min Park[1], Han Gyu Yoon[1], Tae Jung Moon[1], Jun Woo Choi[2], Hee Young Kwon[2*], and Changyeon Won[1*]

[1]***Department of Physics, Kyung Hee University, Seoul 02447, South Korea***

[2]***Center for Spintronics, Korea Institute of Science and Technology, Seoul 02792, South Korea***

*: Corresponding authors

E-mail: soky572@gmail.com, cywon@khu.ac.kr

**Abstract**

This study proposes a novel approach to extract topological properties, specifically the Euler characteristic, from input images using neural networks without relying on large pre-existing datasets but with a single geometric image. Inspired by solid-state physics, where topological properties of magnetic structures are derived from spin field analysis, our model generates a unit vector field from an image, interpreted as a spin configuration. The Euler characteristic is then predicted by computing the skyrmion number of this generated spin configuration. Remarkably, the network learns to construct chiral magnetic textures without access to ground-truth chiral spin configurations, relying instead on only a single, simple geometric image and the straightforward skyrmion number computation. Furthermore, spin configurations generated by independently trained networks can be non-unique due to inherent degrees of freedom. To constrain these degrees of freedom and further refine the spin configuration, we incorporate a magnetic Hamiltonian—comprising exchange interaction, Dzyaloshinskii-Moriya (DM) interaction, and anisotropy—as an additional, physics-informed loss function. We validate the model's efficacy on complex geometrical shapes and demonstrate its applicability to practical tasks.

## 1 Introduction

Topological data analysis (TDA) is a recent and fast-growing field providing a set of new topological and geometric tools to infer relevant features from complex data[1]. It proposes well-founded mathematical theories and computational tools that can be used independently or in combination with other data analysis techniques and methods. Many modern data analysis methods draw inspiration from topological and geometric approaches. It has many significant implications in fields such as materials science[2], and biomedicine[3], where TDA plays a crucial role in characterizing complex structures.

Moreover, to unlock the full potential of TDA, especially when dealing with intricate structures or requiring novel computational paradigms, researchers are increasingly integrating techniques from artificial intelligence (AI)[4]. This synergy enhances tasks like data classification and computational efficiency[5] and often involves drawing inspiration from other scientific domains to develop innovative analysis methods. Recent advancements in medicine have demonstrated the power of integration of AI into topology analysis for detection and classification of abnormal heartbeats[6]. In this approach, TDA helps capture patient-specific variations, enabling the AI model to generalize effectively to unseen individuals.

Another interdisciplinary approach, drawing inspiration directly from condensed matter physics, utilizes the concept of the skyrmion[7–15]. A skyrmion is a topologically protected chiral spin texture that emerges in two-dimensional magnetic systems, and its topological number, called skyrmion number, is computed directly by integrating the solid angles of the skyrmion. Recently, AI-driven methods were proposed to predict the Euler characteristic of images by adapting skyrmion number computation techniques[16,17]. In one approach, the given image is converted into a two-dimensional chiral spin configuration using an image-to-image regression model, followed by the computation of the skyrmion number. The authors confirm that the resulting skyrmion number aligns with the input image's Euler characteristic, which suggests a fundamental connection between the two topological invariants. This approach successfully predicts the Euler characteristic of various complex images, such as indexing Euler characteristic of handwritten digits[18] or counting red blood cells in a microscopic image. The method utilizes numerous chiral spin configurations produced by Monte-Carlo simulation for a training dataset and thus the trained model constructs a specific magnetic texture learned from the given dataset.

Recent advances in machine learning have enabled various approaches for the analysis and prediction of magnetic textures. Supervised learning frameworks have been developed to detect or segment skyrmion-like spin configurations from simulated magnetic images using large, labeled datasets[19]. In addition, physics-informed unsupervised neural networks have been introduced to predict equilibrium spin states from given magnetic parameters, effectively serving as surrogate models for micromagnetic simulations[20]. While these approaches have demonstrated remarkable performance and wide applicability, they inherently depend on extensive labeled datasets or precomputed simulation data.

In this study, we propose a method to acquire topological structures—interpreting spin configurations—and predict Euler characteristic from input images by generating spin configurations, trained directly from a single geometric image rather than a large dataset. A common alternative, supervised learning, typically involves training an image-to-image model using paired input images and target spin configurations with a pixel-to-pixel reconstruction loss. In contrast to such methods, we directly optimize the skyrmion number of the output spin configuration to match the input image's Euler characteristic, without requiring observation of the spin configurations. The neural network learns to construct chiral spin configurations based solely on an input image of a simple geometric shape and the straightforward skyrmion number computation. This approach not only successfully predicts the Euler characteristics of various images but also allows greater flexibility in the resulting spin configuration. Furthermore, we incorporate a magnetic Hamiltonian as an additional loss function to address and control these degrees of freedom, ensuring that the spin configuration is physically stable under the given Hamiltonian. We demonstrate further applications of our approach in analyzing experimental micromagnetic images and counting objects in given images.

## 2 Strategy

The main goal of our study is to predict the Euler characteristic of an input image by computing the skyrmion number of the output spin configuration $\mathbf{S}_{\text{out}}$, which is composed of Heisenberg spins (normalized vectors with three spin components, $S_x$, $S_y$, and $S_z$)[21]. To provide a clearer understanding of the skyrmion number, we present an example in Fig. 1a.

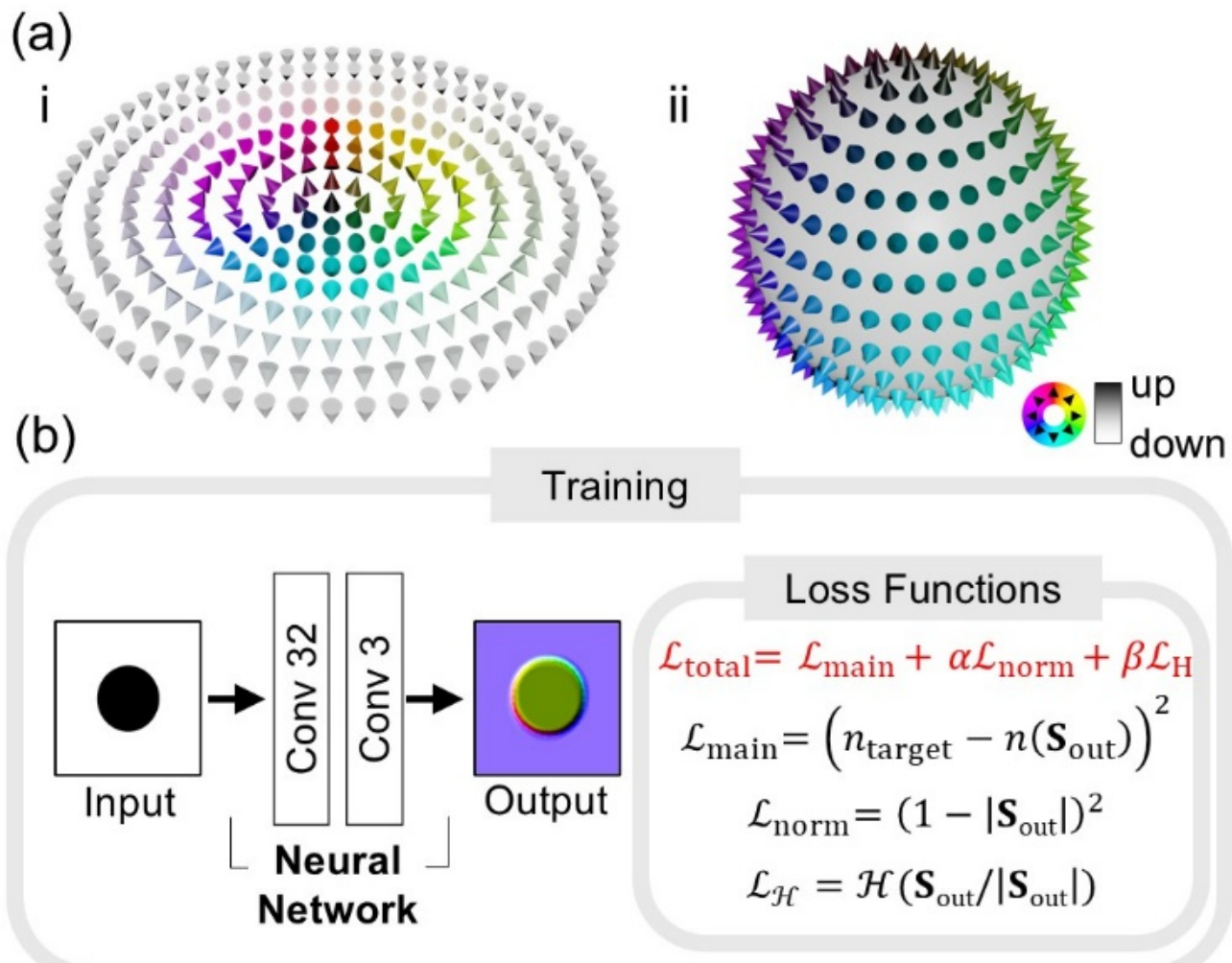

**Figure 1.** (a) Illustrations of a magnetic skyrmion (i) in two-dimensional space and (ii) mapped onto a sphere. The black/white contrast and the color indicate the out-of-plane and the in-plane spin components, respectively. (b) A schematic diagram of the training process of our model. The 'Conv 32' and 'Conv 3' indicate the convolutional neural network layers consisting of 32 and 3 filters, respectively. The input is a one-channel image, representing topological objects. The output of our model is a three-channel image encoding a spin configuration composed of unnormalized spins $\mathbf{S}_{\text{out}}$.

Figure 1a-i shows a magnetic skyrmion with a swirling spin pattern radiating outward from the center, which highlights its characteristic helicity. The skyrmion spin configuration can also be mapped onto a spherical surface, as shown in Fig 1a-ii. It provides a way to visualize the topological nature of skyrmions as they cover the entire sphere. The skyrmion number is determined by integrating the solid angles formed by adjacent spins at each local site, as shown in Eq. 1, yielding $n = 1$ for the magnetic skyrmion.

$$n(\mathbf{S}) = 1 = \frac{1}{4\pi}\int d\Omega = \frac{1}{4\pi}\int \mathbf{S}\cdot\left(\frac{\partial\mathbf{S}}{\partial x}\times\frac{\partial\mathbf{S}}{\partial y}\right)dxdy \quad (1)$$

We construct an image-to-image neural network structure, as illustrated in Fig. 1b, which transforms a one-channel input image into a three-channel output image encoding a spin configuration. The skyrmion number of the resulting spin configuration corresponds to the Euler characteristic of the input image. Notably, our model's sparsity, characterized by its minimal number of trainable parameters, enables training with just a single topological image and its corresponding Euler characteristic as a label.

This method introduces a novel paradigm in which the network autonomously constructs topological information without any observation of the correct solution. While the Euler characteristic of the input image serves as an explicit label, the network internally generates the spin configuration without any ground-truth spin data, guided solely by the topological consistency between the Euler characteristic and the skyrmion number. Despite this limited training data, the model demonstrates exceptional generalization, accurately predicting the Euler characteristics of a variety of complex structures. In this context, the proposed approach exhibits a distinctive characteristic that cannot be categorized as supervised learning though a single labeled data is provided during training. Rather, it is more appropriately described as a self-supervised learning, as the network autonomously learns to construct spin configurations to maintain topological consistency.

In the training stage, the primary objective is to align the skyrmion number of output spin configuration, $n(\mathbf{S}_{\mathrm{out}})$, with a target value, $n_{\mathrm{target}}$, which is chosen as the Euler characteristic of the input image (i.e. $n_{\mathrm{target}} = \chi_{\mathrm{input}}$). As a result, the trained model can predict the Euler characteristic of the input image. This is achieved by minimizing the main loss, $\mathcal{L}_{\mathrm{main}}$, which is the mean squared error between the $n(\mathbf{S}_{\mathrm{out}})$ and $n_{\mathrm{target}}$. As $\mathcal{L}_{\mathrm{main}}$ decreases, the model learns to map input images into chiral spin configurations whose skyrmion number corresponds to the Euler characteristic of the input image. Along with the main loss, two additional losses were incorporated; the normalization loss $\mathcal{L}_{\mathrm{norm}}$, and the Hamiltonian loss $\mathcal{L}_{\mathcal{H}}$. Equation 2 shows the total loss function, $\mathcal{L}_{\mathrm{total}}$, where the $\mathcal{H}$ represents a magnetic Hamiltonian model and the $\alpha$ and $\beta$ denote the coefficients of the normalization loss and the Hamiltonian loss, respectively.

$$\begin{aligned}
\mathcal{L}_{\mathrm{main}} &= \left(n_{\mathrm{target}} - n(\mathbf{S}_{\mathrm{out}})\right)^2 \\
\mathcal{L}_{\mathrm{norm}} &= (1 - |\mathbf{S}_{\mathrm{out}}|)^2 \\
\mathcal{L}_{\mathcal{H}} &= \mathcal{H}(\mathbf{S}_{\mathrm{out}}/|\mathbf{S}_{\mathrm{out}}|) \\
\mathcal{L}_{\mathrm{total}} &= \mathcal{L}_{\mathrm{main}} + \alpha\mathcal{L}_{\mathrm{norm}} + \beta\mathcal{L}_{\mathcal{H}}
\end{aligned} \tag{2}$$

The normalization loss $\mathcal{L}_{\text{norm}}$ is employed to enforce that the norm of the $\mathbf{S}_{\text{out}}$ approaches 1 so that it represents the Heisenberg spin model. Notably, the output spin configuration $\mathbf{S}_{\text{out}}$ has considerable degrees of freedom. Different spin configurations can yield the same skyrmion number, allowing multiple spin configurations to satisfy the ground truth condition of the main loss. Consequently, each training trial may result in different conditions, constructing different patterns of spin configuration. To control the variability in the output spin configuration, we introduce the magnetic Hamiltonian loss, $\mathcal{L}_{\mathcal{H}}$, which evaluates the energetic stability of the output spin configuration based on a given Hamiltonian model $\mathcal{H}$. Practically, we introduce exchange interaction, Dzyaloshinskii-Moriya (DM) interaction, and the out-of-plane anisotropy. Thus, the Hamiltonian model is expressed $\mathcal{H}(\mathbf{S}) = J\left[\left(\frac{\partial \mathbf{S}}{\partial x}\right)^2 + \left(\frac{\partial \mathbf{S}}{\partial y}\right)^2\right] + D\left[\hat{y}\cdot\left(\mathbf{S}\times\frac{\partial \mathbf{S}}{\partial x}\right) - \hat{x}\cdot\left(\mathbf{S}\times\frac{\partial \mathbf{S}}{\partial y}\right)\right] + K[(\mathbf{S}\cdot\hat{z})^2\ ]$, where the $J$, $D$, and the $K$ refer the coefficients of exchange interaction, DM interaction, and the out-of-plane anisotropy[22].

Through this combined strategy of loss functions, our model achieves robust and physically meaningful training outcomes. By balancing the topological accuracy enforced by $\mathcal{L}_{\text{main}}$, the spin normalization constraint $\mathcal{L}_{\text{norm}}$, and the energetic stability imposed by the Hamiltonian loss $\mathcal{L}_{\mathcal{H}}$, we effectively reduce the degrees of freedom in the solution space. Thus, our network is guided towards energetically stable and topologically accurate spin configurations with reduced variability.

## 3 Results

### 3.1 Learning from Single Data

Central to our approach is a training methodology leveraging a fundamental connection between the Euler characteristic $\chi$ and the skyrmion number $n(\mathbf{S})$, which facilitates effective learning from minimal data. We demonstrate this using a cross-validation procedure detailed in Fig. 2. In this procedure, our model undergoes separate training instances, each using only a single image from the dataset, followed by evaluation on the complete dataset. Figure 2a illustrates this setup, where each image in the dataset contains simple shapes, each associated with its Euler characteristic $\chi$. For two-dimensional images, the Euler characteristic represents the number of objects subtracted by the number of holes. Specifically, images featuring single solid shapes (circle, square, and triangle) have $\chi = 1$, while an image containing only a hole (a white circle in a black background) yields $\chi = -1$. An image with two distinct circles has $\chi = 2$, whereas an image of a squared ring results in $\chi =$ 0 (representing one object with one hole, whose contributions cancel). We performed independent training runs, using each of these distinct images (labeled from (i) to (vi) in Figure 2a) individually as the sole training example for a given run.

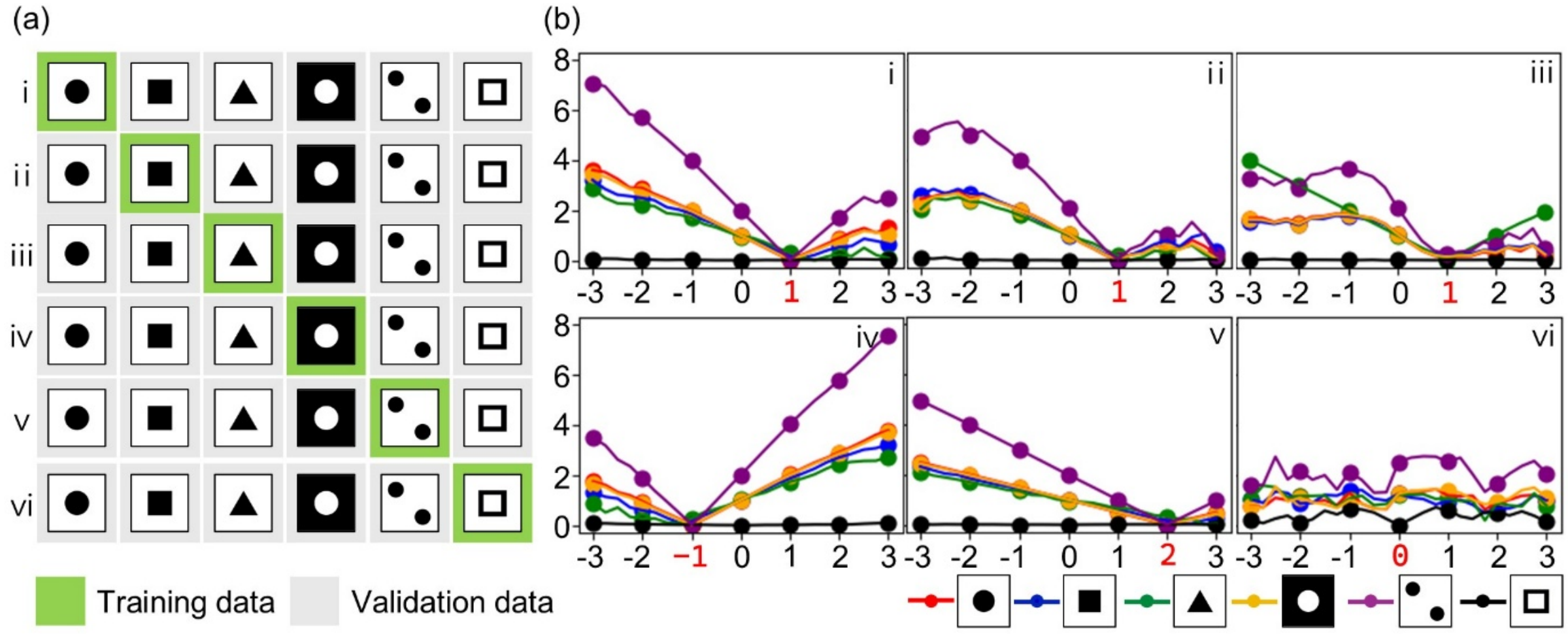

**Figure 2.** (a) Schematic illustration of the cross-validation setting. In each case (i–vi), a single training data (green) is chosen from the dataset. The model is then evaluated on the entire dataset, including the training data and validation data (gray). (b) Cross-validation results shown as graphs corresponding to cases (i–vi) shown in (a). The x-axis represents the targeted skyrmion number, $n_{\text{target}}$. The y-axis represents the absolute error, defined as $|\chi_{\text{test}} - n(\mathbf{S}_{\text{out}})|$, where $\chi_{\text{test}}$ is the input Euler characteristic of the input test image, and $n(\mathbf{S}_{\text{out}})$ is the skyrmion number computed from the model output $\mathbf{S}_{\text{out}}$. Each data point represents the average absolute error calculated over 20 independent trials conducted under identical training conditions.

Figure 2b presents the cross-validation results, showing the absolute error, defined as $|\chi_{\text{test}} - n(\mathbf{S}_{\text{out}})|$, between the input Euler characteristic $\chi_{\text{test}}$ of a test image and the skyrmion number $n(\mathbf{S}_{\text{out}})$ of the model output $\mathbf{S}_{\text{out}}$. This error is plotted against the targeted skyrmion number $n_{\text{target}}$ specified in the main loss $\mathcal{L}_{\text{main}}$ in Eq. 2. The $n_{\text{target}}$ was systematically varied from $-3$ to $3$, including non-integer values. Particularly in cases (i–v), the results demonstrate that when the targeted skyrmion number $n_{\text{target}}$ aligns precisely with the training image's Euler characteristic $\chi_{\text{train}}$, our model accurately learns to generalize this relationship, yielding correct Euler characteristic estimations across entire dataset. Specifically, our model not only maintains the overall structure of the input image but also constructs a spin configuration that yields a skyrmion number that matches

the Euler characteristic of the input image. A detailed explanation and visualization of the output is in the next section.

However, notable exception to this effective learning and generalization are observed in case (vi). An exception occurs when the training image has an Euler characteristic of $\chi_{\mathrm{train}} = 0$, as the case of (vi). In this case, our model fails to learn the underlying topology. Instead, our model constructs a uniform (topologically trivial) spin configuration $\mathbf{S}_{\mathrm{out}}$. While the trivial state correctly yields $n(\mathbf{S}_{\mathrm{out}}) = 0$, matching the training image's Euler characteristic, it fails to capture the non-trivial topology of the ring shape. As a result, our model trained on the ring subsequently performs poorly when evaluated on test images with non-zero $\chi$; it invariably outputs the trivial uniform spin configurations, leading to high errors for these test images. Consequently, training with the ring image and setting $n_{\mathrm{target}} = \chi_{\mathrm{train}} = 0$ does not enable the model to learn the intended topological relationship, resulting instead in the model defaulting to a trivial uniform spin configuration.

For training cases (i-v), the absolute error reaches a minimum across the entire dataset specifically when the target skyrmion number precisely matches the training image's Euler characteristic $\chi_{\mathrm{train}}$, i.e. $n_{\mathrm{target}} = \chi_{\mathrm{train}}$ (indicated by red arrows on the x-axis in each graph). This success demonstrates the model's ability to learn a mapping from an input image to a spin configuration $\mathbf{S}_{\mathrm{out}}$ whose skyrmion number aligns with the training image's Euler characteristic and generalize this mapping effectively across diverse topological images.

For the case of testing the square ring (the black lines in the graphs), the error remains very small regardless of the value of $n_{\mathrm{target}}$. Nevertheless, this result does not directly represent the success of our approach. When computing the skyrmion number of the ring-shaped chiral texture (skyrmionium)[23,24], the solid angle contributions from the inner and outer boundaries cancel each other out. In cases where $n_{\mathrm{target}} \neq \chi_{\mathrm{train}}$, the solid angle on the inner and outer boundaries may be overestimated or underestimated; however, these residuals also cancel each other, resulting in a minimal overall error. Consequently, the small error value for the square ring does not indicate accurate learning of the underlying topology but rather reflects this intrinsic cancellation effect.

We confirm that our model, trained on minimal data, can construct topological spin configuration from input images maintaining its topological properties and Euler characteristics $\chi$ as the outputs skyrmion number $n(\mathbf{S}_{\mathrm{out}})$. Our model learns this task from the fundamental connection between the Euler characteristic and the skyrmion number. Furthermore, it is possible to predict the Euler characteristics of various shapes, even if the shapes are not observed in the training phase.

To further assess the model complexity and the effect of the Hamiltonian loss coefficient ($\beta$), we systematically varied both the number of convolutional filters and the value of $\beta$ in the loss function. In this section, only the exchange interaction is included in the Hamiltonian loss. Detailed results and effects of the Hamiltonian loss will be mainly discussed in part 3.3. For each combination of parameters, 100 independent training trials are conducted, and each trained model is evaluated on ten examples—six sown in Fig. 2 and four further illustrated in Fig. 3.

Table 1 summarizes the proportion of models that successfully generalized the Euler characteristic estimation task when trained from a single circle image. We find that simpler architectures (i.e., with fewer filters) could still achieve successful generalization; however, their success rate was significantly lower than that of larger models. More statistically stable learning dynamics were observed in models with a larger number of trainable parameters, suggesting a reduced probability of convergence to failure modes. Moreover, models trained with nonzero $\beta$ values exhibited higher success ratios across all configurations, indicating that incorporating the Hamiltonian loss term improves training stability and enhances the robustness of learning.

| Filters [a] | Trainable parameters [b] | Training success ratio [c] | | |
|---|---|---|---|---|
| | | $\beta = 0.0$ | $\beta = 0.1$ | $\beta = 1.0$ |
| 1 | 104 | 0.37 | 0.47 | 0.77 |
| 2 | 205 | 0.27 | 0.40 | 0.76 |
| 4 | 407 | 0.31 | 0.42 | 0.84 |
| 8 | 811 | 0.54 | 0.75 | 0.98 |
| 16 | 1619 | 0.75 | 0.95 | 1.00 |
| 32 | 3235 | 0.90 | 1.00 | 1.00 |
| 64 | 6467 | 0.95 | 1.00 | 1.00 |

[a]: The number of filters in the first convolutional layer in the architecture.

[b]: The total number of total trainable parameters including weights and biases.

[c]: The ratio of successful training trials out of 100 runs. Each trial is evaluated on ten examples and is considered successful when the model correctly predicts the Euler characteristic for all examples.

### 3.2 Euler Characteristic Estimation

To demonstrate the efficacy of our model, we apply it to example images including various topological structures, as shown in Fig. 3. We use the model trained on a circular image (see Fig. 1b) with the additional Hamiltonian loss, using parameters as $J = 1.0$, $D = 0.5$, and $K = 0.1$. While the inclusion of the Hamiltonian loss does not affect the Euler number prediction, it restricts the degrees of freedom in the output spin configuration. The non-uniqueness of the ground truth without the Hamiltonian loss, as well as the effect of the Hamiltonian loss, will be discussed in the next section. Here, we focus on validating the model's ability to predict Euler characteristics.

Figure 3(a-c) presents an example application of the model to a solid triangular image. Since a disk and a triangle can be continuously deformed into each other, they are topologically equivalent and share an Euler characteristic of one. The trained model successfully converts the triangular image into a spin configuration, represented as a magnetic skyrmion shaped like the triangle, yielding a skyrmion number of $n = 1$, as shown in Figure 3b. The outer and inner regions of the triangle correspond to the out-of-plane spin components, where the white and black regions denote opposite directions of the magnetic domains. The region between these domains, referred to as the magnetic domain wall, features in-plane spins oriented cyclically. Figure 3c visualizes the local contribution of the solid angle computed from the triangular spin configuration. Due to the uniformity of the magnetic domains, the solid angle density is concentrated in the domain walls, particularly at the vertices where the spin configurations change direction.

Although the skyrmion number computed from the generated spin configuration aligns numerically with the Euler characteristic of the input image, this correspondence does not imply a complete equivalence between the two quantities in a physical sense. In real magnetic systems, topological charge can vary even among configurations that share the same Euler characteristic—for example, skyrmions and magnetic droplets[25,26] may have identical Euler characteristics but distinct topological charges and energetic stability. In our case, the network learns to generate a spin configuration whose skyrmion number matches the Euler characteristic of the input image. As a result, the model spontaneously forms chiral boundary structures without any prior observation of chiral textures or access to pre-existing datasets.

Figure 3d illustrates examples of a solid square, its color-inverted image, and a square ring. These shapes are topologically distinct from each other, resulting in different Euler characteristics. As shown in Fig 3e, the model converts the example images into a magnetic skyrmion, a skyrmion with inverted spins, and a skyrmionium[27] (skyrmion within a skyrmion). The skyrmion has a skyrmion number of $n = 1$, regardless of its specific shape. The skyrmion with inverted spins yields a skyrmion number of $n = -1$ due to the opposite wrapping direction. The skyrmion number can be understood as the product of the winding number at the boundary and the polarity at the skyrmion center. The skyrmion with inverted spins maintains the same winding number (following the clockwise boundary of the skyrmion with a white core, the spins change through the same color order as in the skyrmion with a black core) but has the opposite polarity. The skyrmionium, consisting of a skyrmion with an embedded inverted skyrmion, has a skyrmion number of $n = 0$, as the contributions from the inner and outer regions cancel each other out. These skyrmion numbers align with the Euler characteristics of their respective input images.

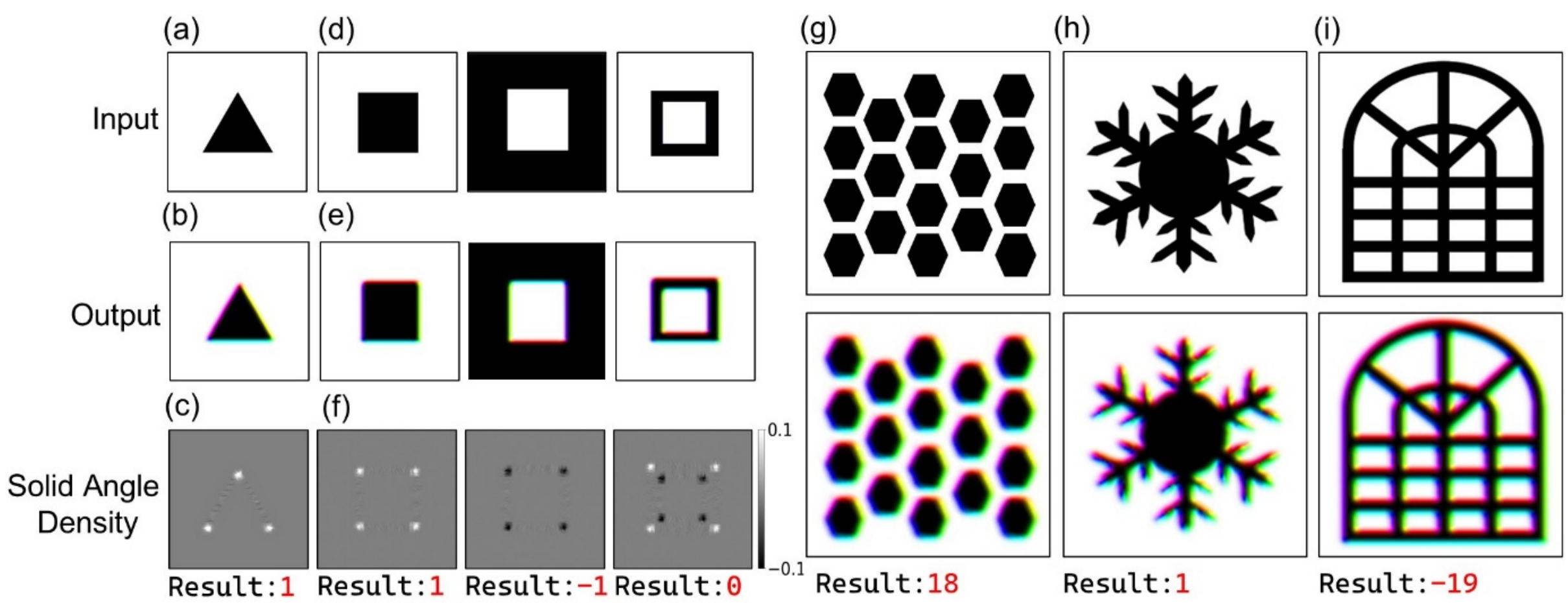

**Figure 3.** Input images and the results of our method. (a) Simple image of a solid triangle (b) The spin configurations converted from (a). (c) The solid angle density calculated from (b). (d) Simple images of a solid square and its color inverted image and square rings. (e) The spin configurations converted from (d). (f) The solid angle density calculated from (e). (g-i) Complex input images and their converted spin configurations for (g) a group of hexagons, (h) a snowflake, and (i) a window frame. The 'Result' denotes the skyrmion number acquired from the spin configurations.

Figure 3f highlights the solid angle densities of the spin configurations in Fig. 3e. Despite the solid square's spin configuration displaying four peaks at its vertices, the total solid angle summation yields a skyrmion number of $n = 1$, consistent with the results for the triangle and circle. This demonstrates that topologically equivalent shapes yield identical skyrmion numbers, validating the model's ability to predict Euler characteristics. The color-inverted image of the solid square generates a spin configuration with opposite spins to the solid square, resulting in $n = -1$. Mathematically, this inversion can be expressed by substituting $\mathbf{S}$ with $-\mathbf{S}$ in the formula $\frac{\Omega}{4\pi} = \frac{1}{4\pi}\mathbf{S} \cdot \left(\frac{\partial \mathbf{S}}{\partial x} \times \frac{\partial S}{\partial y}\right)$, which reverses the solid angle contribution. Furthermore, the square ring features two domain walls with opposing solid angle contribution. The solid angle densities from these two domain walls cancel each other out, leading to a net skyrmion number of $n = 0$, also aligning with the Euler characteristic of the input image.

Given that the Euler characteristic reflects the number of distinct objects in the input image, we evaluated our model on a group of hexagons, as shown in Fig. 3g. Each hexagon contributes a skyrmion number of +1, thus, our method accurately counts the total number of objects as 18. Figure 3h demonstrates further reliability of our method by examining it a complex image, a snowflake with intricate geometrical details. Despite its complexity, the spin configuration of this structure can be continuously deformed into a single skyrmion, yielding a skyrmion number of 1. Furthermore, Fig. 3i presents a window frame along with its corresponding spin configuration. This frame contains 20 holes of varying shapes and sizes. Regardless of the individual shape or size of each hole, each contributes -1 to the total skyrmion number. The outer frame, on the other hand, contributes to +1. As a result, the overall skyrmion number of the window frame is computed as -19. Note that this logic (counting objects as 1 and holes as -1) also roles in identifying the Euler characteristic. All the resulting skyrmion number of our method aligns with the input's Euler characteristics. These examples emphasize the robustness of our model in handling diverse geometries while maintaining accurate Euler characteristic predictions.

Although the proposed model is trained on a single image, the neural network plays a crucial role in learning a continuous mapping between the geometric input and the resulting spin configuration. Unlike conventional image-processing methods such as border-following[28] or morphological filtering, the network internally encodes nonlinear correlations between spatial features and the topological constraints imposed by the physics-informed loss functions. This enables it to generate spin configurations that are not merely geometric transformations of the input but are also consistent with physical quantities such as the skyrmion number and magnetic energy. As a result, the network can generalize to diverse and unseen geometries while preserving topological and energetic consistency.

### 3.3 Effect of Hamiltonian Loss

As previously mentioned, the neural network does not utilize any spin configuration examples or chiral magnetic structures during the training process. Nevertheless, throughout the learning phase, the network automatically learns to form chiral magnetic structures to generate the spin configurations that constitute skyrmions. Notably, different spin configurations can yield the same skyrmion number. Magnetic skyrmions yield the same skyrmion number of $n = 1$, regardless of their specific shapes. Moreover, the background spins can occupy different orientations on the sphere. For instance, an antiskyrmion with a negative winding number, downward core spins, and upward background spins also results in a skyrmion number of $n = 1$. To investigate this non-uniqueness in the output spin configuration, we train our model both with and without incorporating the Hamiltonian loss, which serves to control the non-uniqueness by encouraging energetically stable spin configurations.

Figure 4a shows the independent training results conducted without incorporating the Hamiltonian loss. Each model produces distinct spin configurations while maintaining the same skyrmion number of $n = 1$. Despite these variations, the spin configurations consistently cover the entire sphere surface, preserving the topological characteristics of skyrmions. To further illustrate this, we visualize the spin configurations mapped onto a sphere. Unlike typical skyrmions, the core and background spins are mapped to arbitrary but diametrically opposite points on the sphere (see the clustered points highlighted by red and blue circle). Nevertheless, the overall spin distribution covers the sphere in the same manner as a skyrmion, ensuring the topological equivalence. The trained models used in Fig. 4a successfully predict the Euler characteristics of various input images, such as examples in Fig 4. These results confirm the existence of multiple valid mappings from input images to spin configurations whose skyrmion numbers match the Euler characteristics of the inputs. Training the model without the Hamiltonian loss leads to one of these non-unique mappings.

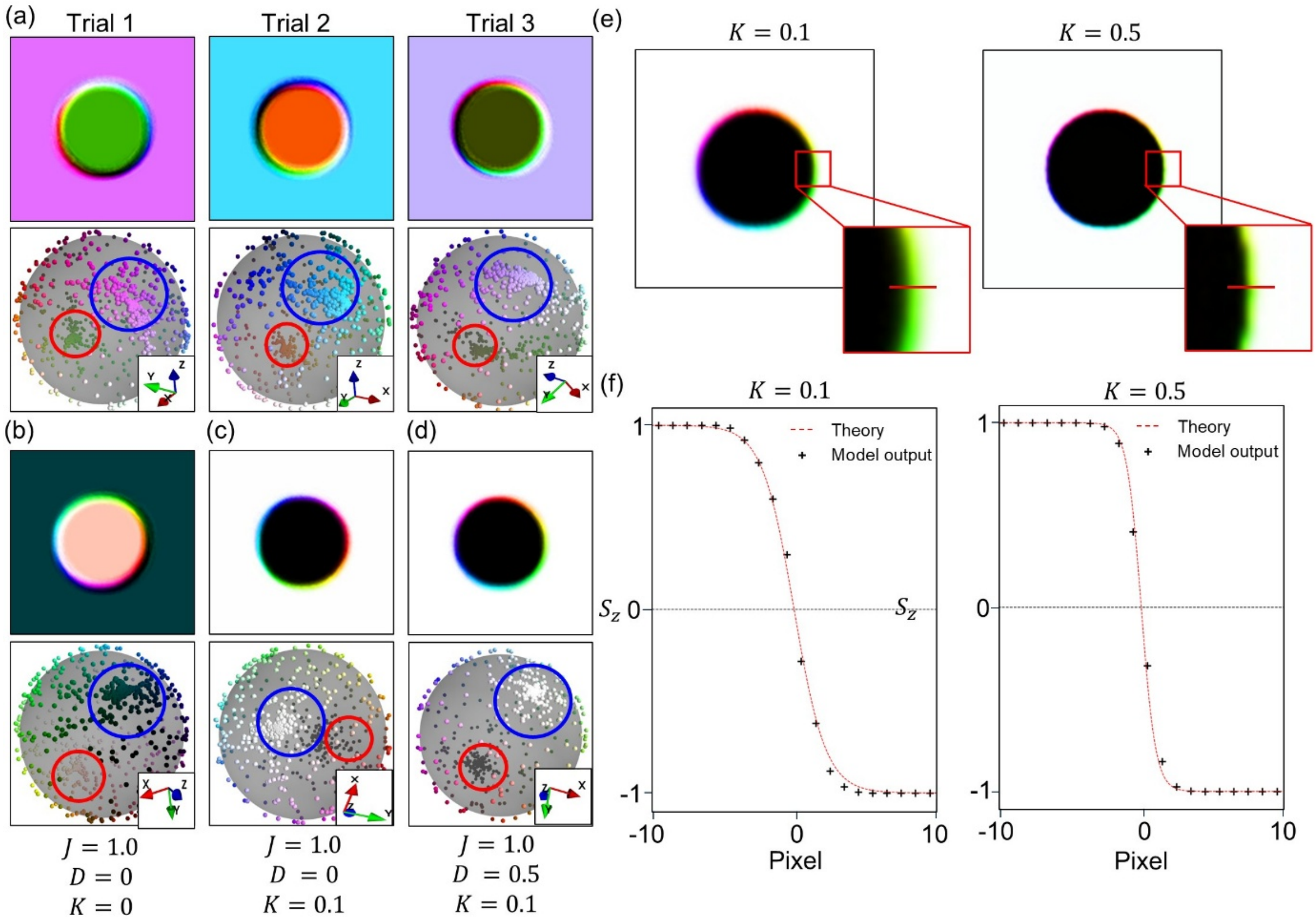

**Figure 4.** Non-uniqueness of the training results of our model and the effect of incorporating the magnetic Hamiltonian loss. (a) Independent training results without incorporating the magnetic Hamiltonian loss. The spin directions are mapped on to a sphere, with the red and blue circles indicating the core and background spins, respectively. The z-axis represents the out-of-plane spin direction, while the x- and y-axes correspond to the in-plane spin directions. (b-d) The training results obtained by incorporating the magnetic Hamiltonian loss with different Hamiltonian parameters: (b) $J = 1.0$, $K = 0.0$, $D = 0.0$, (c) $J = 1.0$, $K = 0.1$, $D = 0.0$, (d) $J = 1.0$, $K = 0.1$, $D = 0.5$. (e) Comparison of training results using different out-of-plane anisotropy values ($K = 0.1$ and $K = 0.5$. (f) Detailed profiles of the model output out-of-plane spin component ($S_z$) along the red lines indicated in (e) . The red dashed lines in the graph represent theoretical $S_z$ profiles corresponding to the chosen anisotropy values. The black/white contrast and the color indicate the out-of-plane and the in-plane spin components, respectively.

To control the inherent non-uniqueness in the model's predictions, we introduce the magnetic Hamiltonian loss, which guides the output toward energetically preferred spin configurations under the given Hamiltonian. Figure 4(b-d) shows the output spin configurations generated by models trained with the magnetic Hamiltonian loss, each using different Hamiltonian parameters. Similar to Fig. 4a, we visualize the spin configurations mapped onto a sphere. Although the exchange interaction contributes to smoother spin variations, the difference between Fig. 4a and Fig. 4b is not prominently visible. However, this term becomes essential for stabilizing the spin texture and preserving its topology when additional types of interactions are introduced. Figure 4c demonstrates the effect of out-of-plane anisotropy on the spin configuration. The anisotropy breaks isotropic symmetry and imposes a preferred spin orientation, favoring spins aligned in the out-of-plane direction. Consequently, the background and core spins are mapped to the out-of-plane directions, in contrast to the arbitrary mapping observed in Fig. 4b. While anisotropy strongly constrains most spins along a preferred axis and significantly reduces the overall non-uniqueness, a certain degree of freedom remains in the domain wall region, where spin components perpendicular to the anisotropy axis are allowed. In this region, the spin texture possesses the degrees of freedom to form cycloidal, or hybrid configurations combining features of both[29]. Finally, Fig. 4d highlights the impact of the Néel-type DM interaction[30,31], which fully eliminates the remaining degrees of freedom in the domain wall region. The DM interaction enforces a cycloidal domain wall pattern[32], producing a typical Néel-type chiral spin configuration, similar to the examples shown in Fig 3.

Furthermore, the magnetic Hamiltonian loss can control the detailed profile of the output spin configuration by testing different out-of-plane anisotropy values. It is well established that, in the two-dimensional magnetic system with exchange interaction and anisotropy, the out-of-plane spin component follows tangent hyperbolic functions with domain wall width $\delta \propto \sqrt{\frac{J}{K}}$. Figure 4e shows the spin configurations obtained from our models trained with the Hamiltonian parameters of $K = 0.1$ and $K = 0.5$. Note that these models have more CNN layers than other models, enhancing their capability of constructing large domain walls. Both models give similar spin configurations but the larger anisotropy results in smaller wall width between the background and the core of the skyrmion. This result aligns with the fact that the domain wall width is inversely proportional to the square root of the anisotropy value. Figure 4f displays the out-of-plane component $S_z$ profiles of the output spin configuration, from the background to the skyrmion core, along with reference line of the theoretical prediction.

We confirm that, by applying Hamiltonian loss, our model successfully constructs skyrmion spin configurations whose detailed structures accurately align with the applied Hamiltonian loss. While a previous study[17] has achieved similar results by training a supervised image-to-image model on an extensive dataset of spin configurations (generated by micromagnetic simulation for a fixed Hamiltonian parameter), our approach offers superior efficiency. It generates physically accurate magnetic structures without the need for a training dataset, instead directly leveraging the Hamiltonian during the training phase. Moreover, this method allows for the generation of arbitrary magnetic textures by simply employing their corresponding Hamiltonian parameters in the training process.

In this study, we adopted a magnetic Hamiltonian that includes exchange interaction, DM interaction, and anisotropy to demonstrate how the Hamiltonian loss can effectively regulate the degrees of freedom in the generated spin configurations. Although other important magnetic interactions, such as dipolar coupling and external magnetic fields, were not considered in the present implementation, these interactions are known to play crucial roles in determining the stability and morphology of magnetic textures in real systems. Our focus here was to examine how the core interactions govern the emergence of topological structures. Nevertheless, the proposed framework is general and can readily incorporate additional physical interactions within the Hamiltonian loss formulation. Incorporating these effects would further enhance the physical realism of the model and facilitate its application to experimentally observed magnetic textures under realistic conditions.

### 3.4 Application

Moreover, we demonstrate applications of our approach on real world images. Figure 5a shows a scanning transmission X-ray microscopy (STXM) image of the magnetic domain structure in in a [Pt(3 nm)/GdFeCo(5 nm)/MgO(1 nm)] multilayer system[33], along with the corresponding spin configurations. The STXM image reveals a stripe domain pattern with alternating upward and downward magnetization directions (dark and bright regions). The brightness of the image represents the out-of-plane component, $S_z$, of local magnetization. The presented spin configurations are constructed from our models trained with different anisotropy values of $0.1$, $0.3$, and $0.5$. As a larger anisotropy value is given, the output spin configuration reveals narrow domain walls. The $S_z$ profile of the domain walls accurately aligns with the theoretical prediction, as shown in Fig. 5b. We confirm that our approach allows us to convert experimental data into the full spin configuration whose detailed structures physically valid for the given Hamiltonian loss.

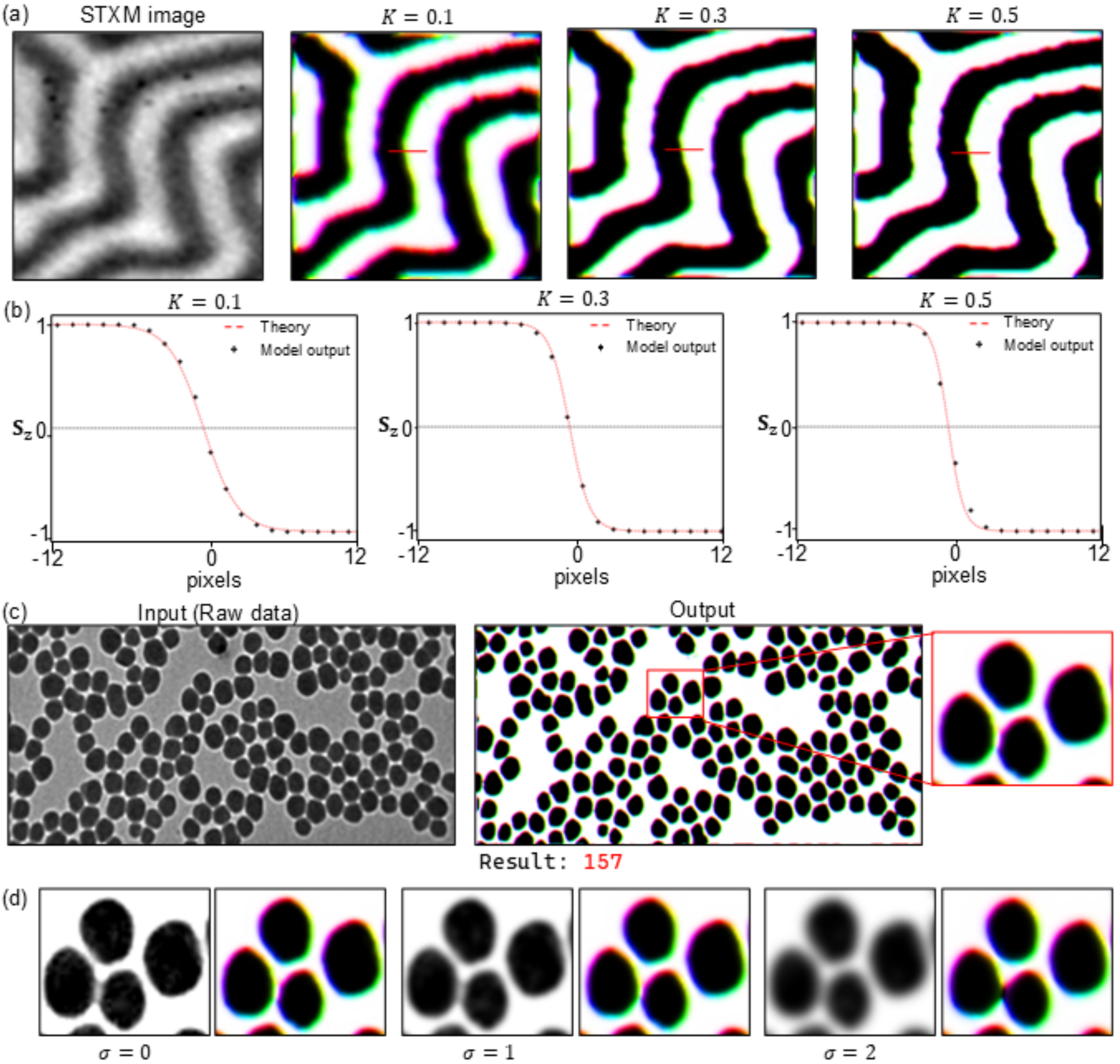

**Figure 5.** (a) Scanning transmission X-ray microscopy image of ferromagnetic system and the corresponding spin configurations of our approach with different anisotropy values of 0.1, 0.3, 0.5. (b) Detailed profiles of the model output out-of-plane spin component ($S_z$) along the red lines indicated in (a). The red dashed lines in the graph represent theoretical references corresponding to the chosen anisotropy values. (c) Silica nanoparticles image and its corresponding spin configuration. *Image by Prof. Mohammed A Sharaf, licensed under Journal of American Science*. (d) Comparison of the neural network outputs under Gaussian blurring with $\sigma = 0$ (no blur), 1, and 2. The first, third and fifth images show the preprocessed input images (brightness rescaled for the neural network input) with different levels of Gaussian bluer, while the second, fourth, and sixth images display their corresponding output spin configurations.

Figure 5c shows another application of our approach, counting silica nanoparticles from a transmission electron microscopy image[34]. Since each nanoparticle is transformed into a skyrmion contributing a skyrmion number of $n = 1$, the number of nanoparticles is inferred from the total skyrmion number in the converted spin configuration. While slight distortions may occur due to the noises or resolution limit, their contribution in the solid angle summation cancels each other out, thus they do not impact the resulting Euler characteristics. As a result, our method computes the skyrmion number of the output spin configuration as $n = 158$, which represents the number of nanoparticles in the input image. We confirm that our method can be utilized for counting the number of objects for a given image.

Despite our approach yielding sufficient results for the examples in Fig 2 and the experimental images in Fig. 5, its performance might be limited when applied to real-world images that include significant noise or blur. Figure 5d demonstrates this limitation by comparing the neural network outputs obtained from inputs with and without Gaussian blur. The inputs without blur and with a weak Gaussian blur ($\sigma = 1$) produce reasonable results, forming four skyrmions corresponding to the four nanoparticles in the input images. In contrast, when input contains after applying a strong Gaussian blur ($\sigma = 2$) Gaussian blur, two of the nanoparticles overlap, and the resulting output spin configuration forms a single, merged skyrmion for the two nanoparticles. Consequently, the resulting Euler characteristic does not correspond to the actual number of nanoparticles. This observation suggests that applying our approach to realistic images requires careful preprocessing, such as amplitude normalization or spatial rescaling, to mitigate the effects of image degradation.

## 4 CONCLUSIONS

In this study, we propose a method that estimates Euler characteristics of images by transforming the images into three-channel spin configurations. Our method accurately predicts Euler characteristics of various geometrically complex images by computing the skyrmion number of output spin configurations. Crucially, a defining strength of our approach is its ability to learn the construction of intricate, even chiral, spin configurations and achieve high predictive accuracy without requiring any pre-collected spin configuration datasets. Instead, the model learns to form these complex magnetic textures guided solely by the topological objective of matching the skyrmion number of its generated spin configuration to the Euler characteristic of a single input geometrical image. The approach also allows for considerable flexibility in the resulting spin configurations. Furthermore, we incorporate the Magnetic Hamiltonian loss to adjust the output spin configuration and control the inherent non-uniqueness in the model's predictions through energetic stabilization. Further applications for real-world images are demonstrated, such as analyzing experimental images of magnetic domain or counting nanoparticles. By bridging machine learning with topological analysis, this study paves the way for advancements in computational physics and materials science.

## ACKNOWLEDGMENT

### Data and Code Availability

All relevant simulation data generated during this study are included within this manuscript. A code implementation of the main methodology of this paper is available at https://github.com/Alexhyungnim/Predicting_Topological_Number_Control_Magnetic_Structures

### Fundings

This research was supported by the National Research Foundation (NRF) of Korea funded by the Korean Government (NRF-2021R1C1C2093113, NRF-2023R1A2C1006050); and by the National Research Council of Science & Technology (No. GTL24041-000).

### Author Contributions.

G.Y. and S.M.P. devised algorithms. G.Y. trained the network and experimented with algorithms. T.J.M., H.G.Y., H.Y.K. contributed to the discussions about the main results. J.W.C. provided experimental image data. H.Y.K. and C.W. equally supervised the work progress. All authors contributed to the final manuscript.

## REFERENCE

1. Chazal, F. & Michel, B. An Introduction to Topological Data Analysis: Fundamental and Practical Aspects for Data Scientists. *Frontiers in Artificial Intelligence* vol. 4 Preprint at https://doi.org/10.3389/frai.2021.667963 (2021).
2. Kramar, M., Goullet, A., Kondic, L. & Mischaikow, K. Persistence of force networks in compressed granular media. *Phys Rev E Stat Nonlin Soft Matter Phys* **87**, (2013).
3. Skaf, Y. & Laubenbacher, R. Topological data analysis in biomedicine: A review. *Journal of Biomedical Informatics* vol. 130 Preprint at https://doi.org/10.1016/j.jbi.2022.104082 (2022).

4. Hensel, F., Moor, M. & Rieck, B. A Survey of Topological Machine Learning Methods. *Frontiers in Artificial Intelligence* vol. 4 Preprint at https://doi.org/10.3389/frai.2021.681108 (2021).

5. Jeong, H. *et al.* A complete Physics-Informed Neural Network-based framework for structural topology optimization. *Comput Methods Appl Mech Eng* **417**, (2023).

6. Dindin, M., Umeda, Y. & Chazal, F. Topological Data Analysis for Arrhythmia Detection Through Modular Neural Networks. in *Lecture Notes in Computer Science (including subseries Lecture Notes in Artificial Intelligence and Lecture Notes in Bioinformatics)* vol. 12109 LNAI (2020).

7. Chen, G. *et al.* Reversible writing/deleting of magnetic skyrmions through hydrogen adsorption/desorption. *Nat Commun* **13**, (2022).

8. Nagaosa, N. & Tokura, Y. Topological properties and dynamics of magnetic skyrmions. *Nature Nanotechnology* vol. 8 Preprint at https://doi.org/10.1038/nnano.2013.243 (2013).

9. Mühlbauer, S. *et al.* Skyrmion lattice in a chiral magnet. *Science (1979)* **323**, (2009).

10. Yu, X. Z. *et al.* Real-space observation of a two-dimensional skyrmion crystal. *Nature* **465**, (2010).

11. Bogdanov, A. N. & Rößler, U. B. Chiral symmetry breaking in magnetic thin films and multilayers. *Phys Rev Lett* **87**, (2001).

12. Liu, J., Shi, M., Mo, P. & Lu, J. Electrical-field-induced magnetic Skyrmion ground state in a two-dimensional chromium tri-iodide ferromagnetic monolayer. *AIP Adv* **8**, (2018).

13. Zhou, Y. *et al.* Dynamically stabilized magnetic skyrmions. *Nat Commun* **6**, (2015).

14. Fert, A., Cros, V. & Sampaio, J. Skyrmions on the track. *Nature Nanotechnology* vol. 8 Preprint at https://doi.org/10.1038/nnano.2013.29 (2013).

15. Jiang, J. *et al.* Current-Controlled Skyrmion Number in Confined Ferromagnetic Nanostripes. *Adv Funct Mater* **33**, (2023).

16. Moon, T. J. *et al.* Computing Euler characteristic of -dimensional objects via a Skyrmion-inspired overlaying (+1)-dimensional chiral field. *Sci Rep* **15**, (2025).

17. Park, S. M., Moon, T. J., Yoon, H. G., Kwon, H. Y. & Won, C. Indexing Topological Numbers on Images by Transferring Chiral Magnetic Textures. *Adv Mater Technol* https://doi.org/10.1002/admt.202400172 (2024) doi:10.1002/admt.202400172.

18. LeCun, Y., Cortes, C. & Burgess, C. J. C. MNIST handwritten digit database. *AT&T Labs [Online]. Available: http://yann. lecun. com/exdb/mnist* **7**, (2010).

19. Labrie-Boulay, I. *et al.* Machine-learning-based detection of spin structures. *Phys Rev Appl* **21**, (2024).

20. Albarracín, F. A. G. Unsupervised machine learning for the detection of exotic phases in skyrmion phase diagrams. http://arxiv.org/abs/2404.10943 (2024).

21. Heisenberg in Leipzig, V. W. *Zur Theorie Des Ferromagnetismus*.

22. Kwon, H. Y. *et al.* Magnetic Hamiltonian parameter estimation using deep learning techniques. *Sci Adv* **6**, (2020).

23. Finazzi, M. *et al.* Laser-induced magnetic nanostructures with tunable topological properties. *Phys Rev Lett* **110**, (2013).

24. Yang, S. *et al.* Reversible conversion between skyrmions and skyrmioniums. *Nat Commun* **14**, (2023).

25. Hoefer, M. A., Silva, T. J. & Keller, M. W. Theory for a dissipative droplet soliton excited by a spin torque nanocontact. *Phys Rev B Condens Matter Mater Phys* **82**, (2010).

26. Mohseni, S. M. *et al.* Spin torque-generated magnetic droplet solitons. *Science (1979)* **339**, 1295–1298

(2013).

27. Ishida, Y. & Kondo, K. Theoretical comparison between skyrmion and skyrmionium motions for spintronics applications. *Jpn J Appl Phys* **59**, (2020).

28. Suzuki, S. & be, K. A. Topological structural analysis of digitized binary images by border following. *Comput Vis Graph Image Process* **30**, (1985).

29. Kao, Y. M., Horng, L. & Cheng, C. H. Analytical studies of the magnetic domain wall structure in the presence of non-uniform exchange bias. *AIP Adv* **11**, (2021).

30. Moriya, T. Anisotropic superexchange interaction and weak ferromagnetism. *Physical Review* **120**, (1960).

31. Dzyaloshinsky, I. A thermodynamic theory of 'weak' ferromagnetism of antiferromagnetics. *Journal of Physics and Chemistry of Solids* **4**, (1958).

32. Chen, G. *et al.* Tailoring the chirality of magnetic domain walls by interface engineering. *Nat Commun* **4**, (2013).

33. Woo, S. *et al.* Current-driven dynamics and inhibition of the skyrmion Hall effect of ferrimagnetic skyrmions in GdFeCo films. *Nat Commun* **9**, (2018).

34. Ibrahim, I. a. M., Zikry, a. a. F. & Sharaf, M. a. Preparation of spherical silica nanoparticles: Stober silica. *Journal of American Science* **6**, (2010).